\documentclass{article}

\usepackage{arxiv}

\usepackage[utf8]{inputenc} 
\usepackage[T1]{fontenc}    
\usepackage{hyperref}       
\usepackage{url}            
\usepackage{booktabs}       
\usepackage{amsfonts}       
\usepackage{nicefrac}       
\usepackage{microtype}      
\usepackage{lipsum}
\usepackage{graphicx}
\graphicspath{ {./images/} }
\usepackage{subcaption}
\usepackage{cases}
\usepackage{amsthm}
\usepackage{amssymb}
\usepackage[ruled]{algorithm2e}
\usepackage{algorithmic}
\usepackage{wrapfig, lipsum}
\usepackage{multirow}
\usepackage[table,xcdraw]{xcolor}
\usepackage[numbers]{natbib}
\usepackage{array}
\usepackage{enumitem}
\usepackage[flushleft]{threeparttable}
\usepackage{tabularx}

\newcolumntype{C}[1]{>{\centering\arraybackslash}m{#1}}

\newcolumntype{R}[1]{>{\raggedleft\arraybackslash}m{#1}}
\newcolumntype{L}[1]{>{\raggedright\arraybackslash}m{#1}}

\newcommand{\customfootnotetext}[2]{{
  \renewcommand{\thefootnote}{#1}
  \footnotetext[0]{#2}}}

\title{Machine Learning-Based COVID-19 Patients Triage Algorithm using Patient-Generated Health Data from Nationwide Multicenter Database}

\author{
 Min Sue Park\textsuperscript{1,*} \\
  \texttt{minsuepark@postech.ac.kr} 
  \And
 Hyeontae Jo\textsuperscript{2,*}\\
    \texttt{jht0116@postech.ac.kr} 
  \And
  Haeun Lee\textsuperscript{3}\\
    \texttt{hlee55313@gmail.com} 
  \And
  Se Young Jung\textsuperscript{3,4,\#}\\
  \texttt{imsyjung@gmail.com} 
   \And
 Hyung Ju Hwang\textsuperscript{1,5,\#} \\
  \texttt{hjhwang@postech.ac.kr}
}

\begin{document}
\maketitle
\customfootnotetext{1}{Department of Mathematics, Pohang University of Science and Technology, Pohang, Republic of Korea}
\customfootnotetext{2}{Basic Science Research Institute, Pohang University of Science and Technology, Pohang, Republic of Korea}
\customfootnotetext{3}{Office of eHealth Research and Business, Seoul National University Bundang Hospital, Seongnam-si, Republic of Korea}
\customfootnotetext{4}{Department of Family Medicine, Seoul National University Bundang Hospital, Seongnam-si, Republic of Korea}
\customfootnotetext{5}{Graduate School of Artificial Intelligence, Pohang University of Science and Technology, Pohang, Republic of Korea}
\customfootnotetext{*}{These authors contributed equally and are co-first authors.}
\customfootnotetext{\#}{Corresponding authors}


\section{Abstract}\label{sec:introduction}

\paragraph{Introduction} A prompt severity assessment model of patients confirmed for having infectious diseases could enable efficient diagnosis while alleviating burden on the medical system. This study aims to develop a SARS-CoV-2 severity assessment model and establish a medical system that allows patients to check the severity of their cases and informs them to visit the appropriate clinic center based on past treatment data of other patients with similar severity levels.
\paragraph{Methods} This paper provides the development processes of a severity assessment model using machine learning techniques and its application on SARS-CoV-2 patients. The proposed model is trained on a nationwide dataset provided by a Korean government agency and only requires patients’ basic personal data, allowing them to judge the severity of their own cases. After modeling, the boosting-based decision tree model was selected as the classifier while mortality rate was interpreted as the probability score. The dataset was collected from all Korean citizens who were confirmed with COVID-19 between February, 2020 and July, 2021 ($N$=149,471).
\paragraph{Results} The experiments achieved high model performance with an approximate precision of 0·923 and Area Under the curve of Receiver Operating Characteristic (AUROC) score of $0$·$950$ [$95$\% Tolerance Interval (TI) $0$·$940$-$0$·$958$, $95$\% Confidence Interval (CI) $0$·$949$-$0$·$950$]. Moreover, our experiments identified the most important variables affecting the severity in the model via sensitivity analysis.
\paragraph{Conclusion} The prompt severity assessment model for managing infectious people has been attained through using a nationwide dataset. It has demonstrated its superior performance by surpassing that of conventional risk assessments. With the model’s high performance and easily accessible features, the triage algorithm is expected to be particularly useful when patients monitor their health status by themselves through smartphone applications.
\paragraph{Keywords} Machine learning, Deep learning, COVID-19, Triage protocol, Mortality, SARS-CoV-2

\noindent\textbf{Key Summary Points} \smallskip\\
\indent \underline{Why carry out this study?}
\begin{itemize}[noitemsep, topsep=0pt, parsep=0pt, partopsep=0pt]
    \item Traditional risk prediction models are limited to identifying the condition of an asymptomatic patient who deteriorates from mild to moderate or extremely severe risk of COVID-19 at triage.
    \item Existing disease risk assessment models were developed with limited size datasets, input variables, and unstandardized independent features without specific ML algorithms. 
\end{itemize}
\indent\indent \underline{What was learned from the study?}
\begin{itemize}[noitemsep, topsep=0pt, parsep=0pt, partopsep=0pt]
    \item This prediction model, trained with patient-generated health data (PGHD) from nationwide COVID-19 screening centers, can be globally utilized to monitor hospitalized or quarantined SARS-CoV-2 confirmed patients daily.
    \item This risk assessment model, developed with multivariable factors like demographic, geographic, and clinical characteristics of a superior performance, can be successfully deployed to manage triage patients with COVID-19. 
\end{itemize}

\section{Introduction}
Countries such as the UK, Singapore, Germany, Portugal, and Israel -- with high vaccination rates -- have created strategies for the new normal after the COVID-19 \cite{1lee_2021,2tham_2021,3more_2021} as many are resuming their pre-COVID-19 lives. However, as the coronavirus mutations cause breakthrough infections, the current vaccine has little effect on reducing the transmission of the virus. The number of confirmed cases in the U.K. and Singapore has been increasing since October, 2021 \cite{4ray_2021}. The variants put a great burden on the healthcare system of those countries \cite{5sim_2021}. Thus, it is evermore imperative to ensure medical readiness at a national level by preparing accurate and reasonable patient severity classification criteria and procedures  \cite{6world2020critical}.

Over the past year and 10 months, S. Korea has experienced four COVID-19 outbreaks, and the occurrence of confirmed cases has been suppressed through the 3T strategy (test, confirmation, investigation, tracking, treatment) and adjustment of social distancing without border blocking and regional blockade \cite{7shin2021national}. According to the Organization for Economic Cooperation and Development, S. Korea has achieved quarantine results without any containment measures, minimizing economic damage, and most effectively blocking the spread of the virus \cite{8allain2020territorial}. Although S. Korea has been relatively performing well in controlling COVID-19, it had difficulty in managing patients whose clinical condition deteriorated from mild to modulate risk level. In fact, there have been cases where patients died at home or a community treatment center, a facility for isolating asymptomatic and mildly symptomatic patients with COVID-19, due to delayed response \cite{9choi2020community, 10arin_2021}.

Thus, a risk prediction model that accurately identifies the condition of a patient who deteriorates from mild to moderate or severe risk is required. Furthermore, it is crucial to triage COVID-19 patients based on the severity of their infection to secure the entire medical system of a nation. For the self-quarantining population of COVID-19, accurate severity-assessment tools are necessary to appraise health status every day \cite{11wang2020role, 12depuydt2021triage}. Several models have been developed to predict the prognosis of confirmed patients or the possibility of COVID-19 diagnosis of patients before confirmation. However, there were several problems: (1) the size of the research datasets was too small, (2) the number of input variables was limited, (3) the non-standard variables were difficult to use by other institutions, or (4) the specific method of using the model was not presented. Moreover, to the best of our knowledge, there was no study on the mortality rate of SARS-CoV-2 according to symptoms at national level while there have been several studies conducted on the establishment of a model for predicting COVID-19 confirmation based on nationwide dataset with features related to COVID-19. Preventing the spread of COVID-19 has difficult aspects such as requiring not only medical staff but also national action. In contrast, lowering the mortality rate can be effectively managed by medical staff by developing an appropriate triage protocol.

Thus, this study aims to review previous research of prediction models for COVID-19 and develop a model predicting mortality rate of SARS-CoV-2 using nationwide multicenter data, thereby allowing patients to easily predict the severity of COVID-19 by entering their Patient-Generated Health Data (PGHD) during quarantine out of hospital.

\section{Methods}\label{sec:results}

\noindent\textbf{Review of previous research} \smallskip\\
\indent The review of previous research was based on a search of three databases: Google Scholar, PubMed, and medRxiv. The following keywords were searched in combination: severity, machine learning, deep learning, COVID-19, triage protocol, mortality, and SARS-CoV-2.

In this paper, we propose a machine learning model that predicts the mortality of SARS-CoV-2 based on questionnaires written by patients.

\noindent\textbf{Data source and study cohort} \smallskip\\
\indent Ranging from February 2020 to July 2021, the dataset was collected by the Korea Disease Control and Prevention Agency (KDCA), a government-affiliated organization, for all Koreans who tested positive for SARS-CoV-2 in Polymerase Chain Reaction (PCR). Our study was approved by the Institutional Review Board of Seoul National University Bundang Hospital (X-2110-717-902). The dataset consists of 149,471 patients who tested positive, of which 2,000 died. The dataset is labeled according to whether the patient is dead or alive, and it is highly imbalanced ($98$·$7$\% imbalance ratio). The features of the dataset are mainly composed of three types of patient data: (i) basic personal information, (ii) types of first symptoms, and (iii) underlying diseases. A detailed description of these features is given in Table \ref{tab:tab1} and Table \ref{tab:tab2}. As mentioned in the introduction, the area of residence is included in the data feature because it affects the degree of virus activation and medicalization scale.

\begin{table}[!htb]\centering
\captionsetup{singlelinecheck= false, format=hang, justification=centering}
\caption{Baseline characteristics of input features.}
\begin{tabular}{|llcc|}
\hline
\multicolumn{1}{|l}{Type}& \multicolumn{1}{l}{Variables}     & \multicolumn{1}{l}{$N$ (Total=149,471)} & \multicolumn{1}{c|}{\%}    \\ \hline
Basic information       & Sex                               &                     &       \\
                        & \hspace{10pt}Male                 & 75,073              & 50·23 \\
                        & \hspace{10pt}Female               & 74,398              & 49·77 \\
                        & Age                               & $\mu$=44·36 ($\sigma$=20·27)       &       \\
                        & Area of residence                 &                     &       \\
                        & \hspace{10pt}Latitude             & $\mu$=36·93 ($\sigma$=0·93)        &       \\
                        & \hspace{10pt}Longitude            & $\mu$=127·39 ($\sigma$=0·76)       &       \\
                        & Body temperature (T, $^\circ$C)              &                     &       \\
                        & \hspace{10pt}T$\leq$36.5     & 121,557             & 81·32 \\
                        & \hspace{10pt}36.5$<$T$<$37.5             & 6,310               & 4·22  \\
                        & \hspace{10pt}37.5$\leq$ T < 38.3  & 17,227              & 11·53 \\
                        & \hspace{10pt}38.3$\leq$ T         & 4,377               & 2·93  \\
Respiratory symptom     & Cough                             &                     &       \\
                        & \hspace{10pt}True                              & 34,201              & 22·88 \\
                        & \hspace{10pt}False                             & 99,997              & 66·90 \\
                        & Sputum                            &                     &       \\
                        & \hspace{10pt}True                              & 17,108              & 11·45 \\
                        & \hspace{10pt}False                             & 117,090             & 78·34 \\
                        & Sore throat                       &                     &       \\
                        & \hspace{10pt}True                              & 25,078              & 16·78 \\
                        & \hspace{10pt}False                             & 109,120             & 73·00 \\
                        & Dyspnea                           &                     &       \\
                        & \hspace{10pt}True                              & 1,962               & 1·31  \\
                        & \hspace{10pt}False                             & 132,236             & 88·47 \\
Non-respiratory symptom & Musculoskeletal pain              &                     &       \\
                        & \hspace{10pt}True                              & 24,017              & 16·07 \\
                        & \hspace{10pt}False                             & 110,181             & 73·71 \\
                        & Headache                          &                     &       \\
                        & \hspace{10pt}True                              & 16,337              & 10·93 \\
                        & \hspace{10pt}False                             & 117,861             & 78·85 \\
                        & Chill                             &                     &       \\
                        & \hspace{10pt}True                              & 17,227              & 11·53 \\
                        & \hspace{10pt}False                             & 116,971             & 78·26 \\
                        & Ageusia                           &                     &       \\
                        & \hspace{10pt}True                              & 4,846               & 3·24  \\
                        & \hspace{10pt}False                             & 129,352             & 86·54 \\
                        & Anosmia                           &                     &       \\
                        & \hspace{10pt}True                              & 5,498               & 3·68  \\
                        & \hspace{10pt}False                             & 128,700             & 86·10 \\ \hline
\end{tabular}
\label{tab:tab1}
\end{table}

\begin{figure}[!htb]
    \centering
    \includegraphics[width=.95\textwidth]{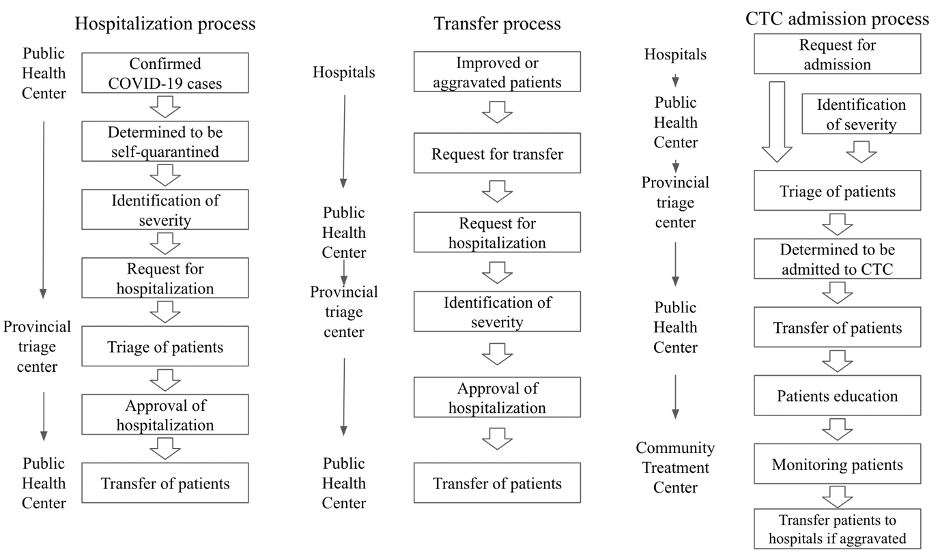}
    \caption{Management strategy of COVID-19 confirmed cases in S.Korea.}
    \label{fig:fig1}
\end{figure}

The data was collected from 1,382 designated COVID-19 screening centers in S. Korea. These centers consist of National Safe Hospitals (263), Dedicated Respiratory Clinics (518), Screening clinics in public health centers (627), Temporary Screening Office (200), and Car Mobile Screening Clinics (15). The process of initial screening, transfer, admission to a hospital or community treatment center (CTC) is presented in Figure \ref{fig:fig1}.

\begin{table}[!htb]\centering
\begin{threeparttable}
\captionsetup{singlelinecheck= false, format=hang, justification=centering}
\caption{Underlying diseases of study participants.}
\begin{tabular}{|lllllllclllllclllclll|}
\hline
\multicolumn{1}{|c}{} & \multicolumn{3}{c}{\multirow{2}{*}{Disease}}  &  &  & \multicolumn{4}{c}{\multirow{2}{*}{Count}}   & \multicolumn{11}{c|}{\, Total ($N$=149,471)}                                                                                                \\
                      & \multicolumn{3}{c}{}                          &  &  & \multicolumn{4}{c}{}                         & \multicolumn{1}{c}{} &  &  & $N$       & \multicolumn{1}{c}{} & \multicolumn{1}{c}{} &  & \%    & \multicolumn{1}{c}{} &  &  \\ \hline
                      &  &  & \multirow{4}{*}{Liver disease}          &  &  &           & \multicolumn{2}{c}{0} &          &                      &  &  & 148,632 &                      &                      &  & 99·44 &                      &  &  \\
                      &  &  &                                         &  &  &           & \multicolumn{2}{c}{1} &          &                      &  &  & 354     &                      &                      &  & 0·24  &                      &  &  \\
                      &  &  &                                         &  &  &           & \multicolumn{2}{c}{2} &          &                      &  &  & 475     &                      &                      &  & 0·32  &                      &  &  \\
                      &  &  &                                         &  &  &           & \multicolumn{2}{c}{3} &          &                      &  &  & 10      &                      &                      &  & 0·01  &                      &  &  \\ \hline
                      &  &  & \multirow{6}{*}{Cancer}                 &  &  &           & \multicolumn{2}{c}{0} &          &                      &  &  & 147,260 &                      &                      &  & 98·52 &                      &  &  \\
                      &  &  &                                         &  &  &           & \multicolumn{2}{c}{1} &          &                      &  &  & 594     &                      &                      &  & 0·4   &                      &  &  \\
                      &  &  &                                         &  &  &           & \multicolumn{2}{c}{2} &          &                      &  &  & 1,423   &                      &                      &  & 0·95  &                      &  &  \\
                      &  &  &                                         &  &  &           & \multicolumn{2}{c}{3} &          &                      &  &  & 187     &                      &                      &  & 0·13  &                      &  &  \\
                      &  &  &                                         &  &  &           & \multicolumn{2}{c}{4} &          &                      &  &  & 5       &                      &                      &  & 0·00  &                      &  &  \\
                      &  &  &                                         &  &  &           & \multicolumn{2}{c}{5} &          &                      &  &  & 2       &                      &                      &  & 0·00  &                      &  &  \\ \hline
                      &  &  & \multirow{2}{*}{Diabetes mellitus}      &  &  &           & \multicolumn{2}{c}{0} &          &                      &  &  & 139,063 &                      &                      &  & 93·04 &                      &  &  \\
                      &  &  &                                         &  &  &           & \multicolumn{2}{c}{1} &          &                      &  &  & 10,408  &                      &                      &  & 6·96  &                      &  &  \\ \hline
                      &  &  & \multirow{6}{*}{Cardiovascular disease} &  &  &           & \multicolumn{2}{c}{0} &          &                      &  &  & 127,608 &                      &                      &  & 85·37 &                      &  &  \\
                      &  &  &                                         &  &  &           & \multicolumn{2}{c}{1} &          &                      &  &  & 2,165   &                      &                      &  & 1·45  &                      &  &  \\
                      &  &  &                                         &  &  &           & \multicolumn{2}{c}{2} &          &                      &  &  & 18,719  &                      &                      &  & 12·52 &                      &  &  \\
                      &  &  &                                         &  &  &           & \multicolumn{2}{c}{3} &          &                      &  &  & 825     &                      &                      &  & 0·55  &                      &  &  \\
                      &  &  &                                         &  &  &           & \multicolumn{2}{c}{4} &          &                      &  &  & 139     &                      &                      &  & 0·09  &                      &  &  \\
                      &  &  &                                         &  &  &           & \multicolumn{2}{c}{5} &          &                      &  &  & 15      &                      &                      &  & 0·01  &                      &  &  \\ \hline
                      &  &  & \multirow{3}{*}{Renal disease}          &  &  &           & \multicolumn{2}{c}{0} &          &                      &  &  & 148,698 &                      &                      &  & 99·48 &                      &  &  \\
                      &  &  &                                         &  &  &           & \multicolumn{2}{c}{1} &          &                      &  &  & 758     &                      &                      &  & 0·51  &                      &  &  \\
                      &  &  &                                         &  &  &           & \multicolumn{2}{c}{2} &          &                      &  &  & 15      &                      &                      &  & 0·01  &                      &  &  \\ \hline
                      &  &  & \multirow{4}{*}{Degenerative disease}   &  &  &           & \multicolumn{2}{c}{0} &          &                      &  &  & 146,945 &                      &                      &  & 98·31 &                      &  &  \\
                      &  &  &                                         &  &  &           & \multicolumn{2}{c}{1} &          &                      &  &  & 2,331   &                      &                      &  & 1·56  &                      &  &  \\
                      &  &  &                                         &  &  &           & \multicolumn{2}{c}{2} &          &                      &  &  & 193     &                      &                      &  & 0·13  &                      &  &  \\
                      &  &  &                                         &  &  &           & \multicolumn{2}{c}{3} &          &                      &  &  & 2       &                      &                      &  & 0·00  &                      &  &  \\ \hline
                      &  &  & \multirow{4}{*}{Lung disease}           &  &  &           & \multicolumn{2}{c}{0} &          &                      &  &  & 147,253 &                      &                      &  & 98·52 &                      &  &  \\
                      &  &  &                                         &  &  &           & \multicolumn{2}{c}{1} &          &                      &  &  & 2,086   &                      &                      &  & 1·40  &                      &  &  \\
                      &  &  &                                         &  &  &           & \multicolumn{2}{c}{2} &          &                      &  &  & 122     &                      &                      &  & 0·08  &                      &  &  \\
                      &  &  &                                         &  &  &           & \multicolumn{2}{c}{3} &          &                      &  &  & 10      &                      &                      &  & 0·01  &                      &  &  \\ \hline
\end{tabular}
\label{tab:tab2}
\begin{tablenotes}
\small
\item$\cdot$\textbf{Liver disease} includes hepatitis B, cirrhosis, and any other hepatitis.
\item$\cdot$\textbf{Cancer} includes liver cancer, thyroid cancer, oral cancer, acute myelogenous white blood, ovarian cancer, brain cancer, colon cancer, lymphoma, chronic myelogenous white blood, bladder cancer, esophageal cancer, cancer, stomach cancer, cervical cancer, uterine cancer, prostate cancer, rectal cancer, skin cancer, hematoma, laryngeal cancer, prostate cancer, hematologic cancer, hematoma, and blood cancer.
\item$\cdot$\textbf{Cardio-cerebrovascular disease} includes hypertension, stroke, cerebral infarction, myocardial infarction, myocardial hemorrhage, arteriosclerosis, and angina.
\item$\cdot$\textbf{Renal disease} includes renal failure, renal failure, and glomerular disease.
\item$\cdot$\textbf{Degenerative disease} includes Alzheimer disease, other dementia, and Parkinson disease.
\item$\cdot$\textbf{Lung disease} includes emphysema and any other lung disease.
\end{tablenotes}
\end{threeparttable}
\end{table}

The triage process of COVID-19 confirmed patients was initiated based on the severity of their symptoms: asymptomatic to mild, moderate, severe, and critical. Symptoms were assessed by telephone interviews or face-to-face in the first-visit facility, and patients were quarantined at designated facilities according to their severity. Asymptomatic and mildly symptomatic patients were admitted to CTCs. Meanwhile, patients with an aggravated severity were hospitalized at tertiary hospitals. The referral system at each level of medical care aims to allow for patients to be efficiently transferred to a higher level of care before worsening clinical status \cite{38kim2020south}.

The overall process of hospitalization and transfer is presented in Figure \ref{fig:fig1}.

\noindent\textbf{Data collection and measurement} \smallskip\\
\indent Previous studies revealed that the outbreaks of COVID-19 were associated with latitude, temperature, and humidity measurements, which reflects seasonal variation in the incidence of respiratory viruses \cite{39sajadi2020temperature,40tian2020distinct}. Thus, geographic information of latitude and longitude have been integrated into our model.
 
Easy-to-measure features are defined as variables such as body temperature, pulse rate, respiratory rate, blood pressure, any symptoms, and past medical history that can be directly collected from patients without much delay. 

\noindent\textbf{Outcome definition} \smallskip\\
\indent The outcome was defined as deceased cases due to COVID-19 in hospitals, CTCs, and at homes. The mortality cases were collected by the KDCA from National Statistics.

\noindent\textbf{Feature generation} \smallskip\\
\indent We observed that the structural stability of individual SARS-CoV-2 could be affected by the temperature and humidity of the atmosphere \cite{41sharma2021structural}. In addition, hospitalization rates may vary depending on access to medical resources and the severity of previous diseases \cite{42riley2012health}. For these reasons, we utilized additional features such as the date of the onset of symptoms (in months), the area of residence (in longitude and latitude coordinates), and underlying patient symptoms. 

The features of the dataset provided by KDCA are listed as follows: sex, age, body temperature, clinical symptoms (cough, sputum, sore throat, dyspnea, musculoskeletal pain, headache, chill, ageusia, anosmia), self-reported underlying diseases. For body temperature (T), we divided patients and categorized them into 4 subgroups: (1) no fever with T$\leq$ $36$·$5$ $^\circ$C, (2) mild elevation of body temperature with $36$·$5$ $^\circ$C $<$ T $<$ $37$·$5$ $^\circ$ C, (3) mild fever with $37$·$5$ $^\circ$ C $\leq$ T $<$ $38$·$3$ $^\circ$ C, and (4) overt fever with $38$·$3$  $^\circ$ C $\leq$ T. All clinical symptoms have binary values; true or false. Since underlying diseases are self-reported in a free format, we manually classified the reported diseases into 7 subgroups: liver disease, cancer, diabetes mellitus, cardio-cerebrovascular disease, renal disease, degenerative disease, and lung disease. Thus, if a patient had lung cancer and liver cancer, they were assigned a value of 2 to the feature named `cancer' for this patient. This was done to reduce the sparsity of our dataset. Since there are so many different diseases, our dataset would become very sparse if we treated each disease as a different feature. If a model is naively trained on a given sparse dataset, the performance of the model would degenerate; and worse, it could increase the chances of the model wrongly predicting the mortality probability for a patient with a rare disease. Moreover, requiring many features would lower user convenience.

\noindent\textbf{Training and evaluation} \smallskip\\
\indent We split the dataset into training sets and test sets with an 80:20 ratio, and the model was evaluated on the test set. We used a tree-based gradient boosting machine learning model with binary logistic objectives, XGBoost (XGB) \cite{43chen2016xgboost}. This model is a decision-tree-based ensemble machine learning model known for its powerful performance in classification problems in various fields 
\cite{park2021explainability,hwang2020hybrid}. Since this is a tree-based model, it has the advantage of being able to process data with missing values \cite{44josse2019consistency}. Another benefit of using gradient boosting algorithms is that they enable straightforward measurement of feature importance scores in prediction by calculating how useful each feature is in the construction of the weak learners within the model. Therefore, this method does not tell us how positively or negatively the features affected the prediction and does not consider the association relations among features in making predictions.

Meanwhile, originating from game theory, the SHapley Additive exPlanations (SHAP) algorithm \cite{45lundberg2017unified} is used to compute Shapley values \cite{46lipovetsky2001analysis} for each feature, where each Shapley value represents the impact of the feature to which it is associated and predicted. When used for tree-based models, SHAP has the great advantage of being able to calculate Shapley values relatively quickly. Therefore, we have utilized it to identify the principal features in model prediction.

The model was evaluated on the test set using various metrics, including Area Under the curve of Receiver Operating Characteristic (AUROC), Area Under the Precision-Recall Curve (AUPRC), F1-score, precision, sensitivity, and specificity. Moreover, we performed a decision curve analysis on the model. ROC analysis provides information about diagnostic test performance; a ROC curve consists of the True-Positive (TP) and False-Positive (FP) rate and demonstrates the discriminatory ability of a binary classifier system by varying the discriminant thresholds. In other words, the discriminatory ability of the test could be powerful when the vertex of the curve is closer to the upper left (high TP rate and low FP rate). In addition, the baseline for AUROC is always $0$·$5$.

On the other hand, PR curves plot the precision against the recall, and AUPRC is especially useful for imbalanced data in a setting where we focus more on detecting the positive examples. Unlike AUROC, the baseline for AUPRC is equal to the fraction of positives. This means that obtaining an AUPRC of $0$·$4$ on a class with $10$\% positives is good but obtaining an AUPRC of $0$·$6$ on a class with $80$\% positives is undesirable \cite{47beger2016precision}.

\section{Results}

\begin{figure}[!htb]
    \centering
    \includegraphics[width=.95\textwidth]{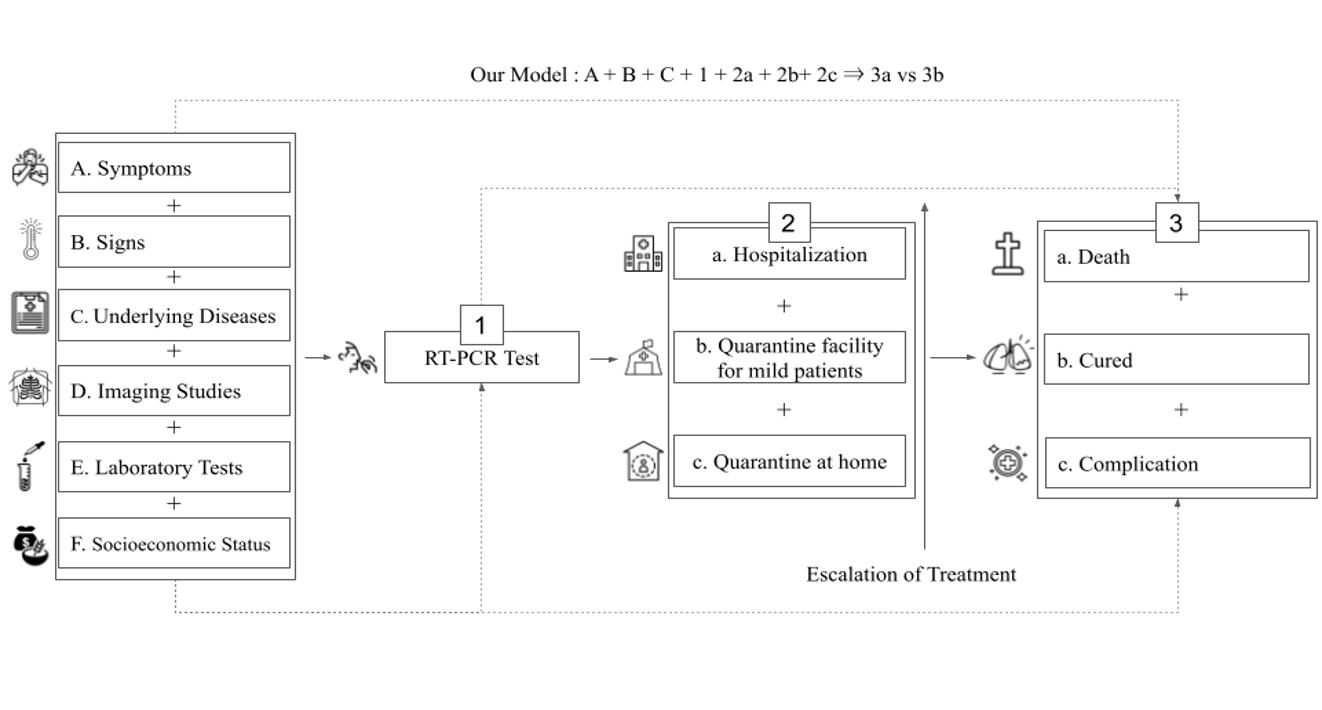}
    \caption{Classification of the previous prediction models according to the type of learning data and type of prediction models.}
    \label{fig:fig2}
\end{figure}

\noindent\textbf{Literature Review} \smallskip\\
\indent Previous research was classified according to the five classification criteria: (1) type of learning data, (2) type of prediction models, (3) outcome variables, (4) data type, and (5) whether or not easy-to-measure input features were utilized. In terms of modeling and utilizing the prediction models, they have four major components: gathering patients’ information such as symptoms, signs, previous medical history; results of imaging studies; and laboratory tests; confirmation of COVID-19 through Reverse Transcriptase Polymerase Chain Reaction (RT-PCR) test; and triage of confirmed cases. The schematic flow of management for COVID-19 patients is presented in Figure \ref{fig:fig2}.

In terms of outcome variables, previous studies were classified into four major classes. 

\begin{itemize}
    \item Outcome Class 1: Diagnosis \\
        \indent\quad\quad A + B $\Rightarrow$ 1 (\citet{13zoabi2021machine, 20menni2020real})\\
        \indent\quad\quad B $\Rightarrow$ 1 (\citet{14yanamala2021vital}) \\
        \indent\quad\quad D $\Rightarrow$ 1 (\citet{15gozes2020rapid,16song2021deep,18jin2020development,19punn2021automated}) \\
        \indent\quad\quad A + B + C + E + 2a $\Rightarrow$ 1 (\citet{17feng2021novel})
        
    \item Outcome Class 2: Mortality \\
        \indent\quad\quad F + 1 + 2a + 2b + 2c $\Rightarrow$ 3a vs 3b (\citet{21cifuentes2021socioeconomic}) \\
        \indent\quad\quad A + B + C + E + 1 + 2a $\Rightarrow$ 3a vs 3b (\citet{24her2021clinical})\\
        \indent\quad\quad C + 1 + 2a + 2b + 2c $\Rightarrow$ 3a vs 3b (\citet{22cho2021impact})\\
        \indent\quad\quad C + E + 1 + 2a $\Rightarrow$ 3a vs 3b (\citet{23ikemura2021using})
    
    \item Outcome Class 3: Mortality and complication \\
        \indent\quad\quad B + D + E + 1 + 2a $\Rightarrow$ (3a + 3c) vs 3b (\citet{26shamout2021artificial}) \\
        \indent\quad\quad C + E + 1 + 2a $\Rightarrow$ (3a + 3c) vs 3b (\citet{25subudhi2021comparing})\\
        \indent\quad\quad A + B + C + E + 1 + 2a $\Rightarrow$ (3a + 3c) vs 3b (\citet{27marcos2021development})\\
        \indent\quad\quad A + B + C + 1 + 2a $\Rightarrow$ (3a + 3c) vs 3b (\citet{28kim2020easy})\\
        \indent\quad\quad C + E + 1 + 2a $\Rightarrow$ (3a + 3c) vs 3b (\citet{29su2021clinical})
    
    \item Outcome Class 4 : Complication\\
        \indent\quad\quad A + B + C + 1 + 2a + 2b + 2c $\Rightarrow$ 3b vs 3c (\citet{30rinderknecht2021predicting}) \\
        \indent\quad\quad A + B + C + D + E + 1 + 2a $\Rightarrow$ 3b vs 3c (\citet{31wang2021icovid})
\end{itemize}

We reviewed 19 previous research and classified them by the four classification criteria and four major outcome classes.  The result is presented in Table \ref{tab:tab3}.

\begin{table}[!htb]\centering
\begin{threeparttable}
\captionsetup{singlelinecheck= false, format=hang, justification=centering}
\caption{Previous research regarding COVID-19 prediction models.}
\begin{tabular}{|C{3cm}L{3cm}|C{1.5cm}|C{2cm}|C{1.7cm}|C{1.5cm}|C{2.5cm}|}
\hline
\multicolumn{1}{|c|}{Class}               & \multicolumn{1}{c|}{Studies} & \begin{tabular}[C{0cm}]{@{}c@{}}Prediction\\ type\end{tabular} & \begin{tabular}[C{0cm}]{@{}c@{}}Outcome\\ variable\end{tabular}          & \begin{tabular}[C{0cm}]{@{}c@{}}Data type\end{tabular}  & \begin{tabular}[c]{@{}c@{}}Sample\\ size\end{tabular}                      & \begin{tabular}[c]{@{}c@{}}Easy-to-measure\\ input features\end{tabular} \\ \hline
\multicolumn{2}{|c|}{Our model}                                          & Prognosis       & Mortality                                                           & Nationwide & 149,471                          & Yes                                              \\ \hline
\multicolumn{1}{|C{0cm}|}{}                    & \citet{13zoabi2021machine}                           & Diagnosis       & RT-PCR\textsuperscript{$\dagger$}                                                              & Nationwide & 99,232    & Yes                                               \\ \cline{2-7} 
\multicolumn{1}{|C{0cm}|}{}                    & \citet{14yanamala2021vital}                           & Diagnosis       & RT-PCR                                                              & Local      & 3,883                            & No                                                                       \\ \cline{2-7} 
\multicolumn{1}{|C{0cm}|}{}                    & \citet{15gozes2020rapid}                           & Diagnosis       & RT-PCR                                                              & Local      & 157       & No                                                \\ \cline{2-7} 
\multicolumn{1}{|c|}{}                    & \citet{16song2021deep}                           & Diagnosis       & RT-PCR                                                              & Local      & 275       & No                                                \\ \cline{2-7} 
\multicolumn{1}{|c|}{}                    & \citet{17feng2021novel}                           & Diagnosis       & RT-PCR                                                              & Local      & 164                              & No                                                                       \\ \cline{2-7} 
\multicolumn{1}{|c|}{}                    & \citet{18jin2020development}                           & Diagnosis       & RT-PCR                                                              & Local      & 11,356    & No                                                \\ \cline{2-7} 
\multicolumn{1}{|c|}{}                    & \citet{19punn2021automated}                           & Diagnosis       & RT-PCR                                                              & Local      & 1,214     & No                                                \\ \cline{2-7} 
\multicolumn{1}{|c|}{\multirow{-8}{*}{1}} & \citet{20menni2020real}                           & Diagnosis       & RT-PCR                                                              & Nationwide & 2,618,862 & Yes                                               \\ \hline
\multicolumn{1}{|c|}{}                    & \citet{21cifuentes2021socioeconomic}                           & Prognosis       & Mortality                                                           & Nationwide & 1,033,218 & Yes                                               \\ \cline{2-7} 
\multicolumn{1}{|c|}{}                    & \citet{22cho2021impact}                           & Prognosis       & Mortality                                                           & Nationwide & 7,590     & No                                                \\ \cline{2-7} 
\multicolumn{1}{|c|}{}                    & \citet{23ikemura2021using}                           & Prognosis       & Mortality                                                           & Local      & 4,313     & No                                                \\ \cline{2-7} 
\multicolumn{1}{|c|}{\multirow{-4}{*}{2}} & \citet{24her2021clinical}                           & Prognosis       & Mortality                                                           & Nationwide & 5,628                            & No                                                                       \\ \hline
\multicolumn{1}{|c|}{}                    & \citet{25subudhi2021comparing}                           & Prognosis       & Complication or Mortality &  Local      & 10,826    & No                                                \\ \cline{2-7} 
\multicolumn{1}{|C{0cm}|}{}                    & \citet{26shamout2021artificial}                           & Prognosis       & Complication or Mortality & Local      & 3,661     & No                                                \\ \cline{2-7} 
\multicolumn{1}{|C{0cm}|}{}                    & \citet{27marcos2021development}                           & Prognosis       & Complication or Mortality & Local      & 1,270                            & No                                                                       \\ \cline{2-7} 
\multicolumn{1}{|C{0cm}|}{}                    & \citet{28kim2020easy}                           & Prognosis       & Complication or Mortality & Nationwide & 4,787     & Yes                                               \\ \cline{2-7} 
\multicolumn{1}{|c|}{\multirow{-8.5}{*}{3}} & \citet{29su2021clinical}                           & Prognosis       & Complication or Mortality & Local      & 14,418    & No                                                \\ \hline
\multicolumn{1}{|c|}{}                    & \citet{30rinderknecht2021predicting}                           & Prognosis       & Complication                                                        & Nationwide & 15,753    & Yes                                               \\ \cline{2-7} 
\multicolumn{1}{|c|}{\multirow{-3}{*}{4}} & \citet{31wang2021icovid}                           & Prognosis       & Complication                                                        & Local      & 3,008     & No                                                \\ \hline
\end{tabular}\label{tab:tab3}
\begin{tablenotes}
\small
\item $\cdot$ RT-PCR: Reverse Transcription Polymerase Chain Reaction
\end{tablenotes}
\end{threeparttable}
\end{table}

The baseline characteristics of the input features used in the research are presented in Table \ref{tab:tab1} and Table \ref{tab:tab2}. The area of residence for each confirmed patient has been converted to floating-point variables using the Python Google Maps API client due to its large scale.

\begin{figure}[!htb]
    \centering
    \includegraphics[width=.8\textwidth]{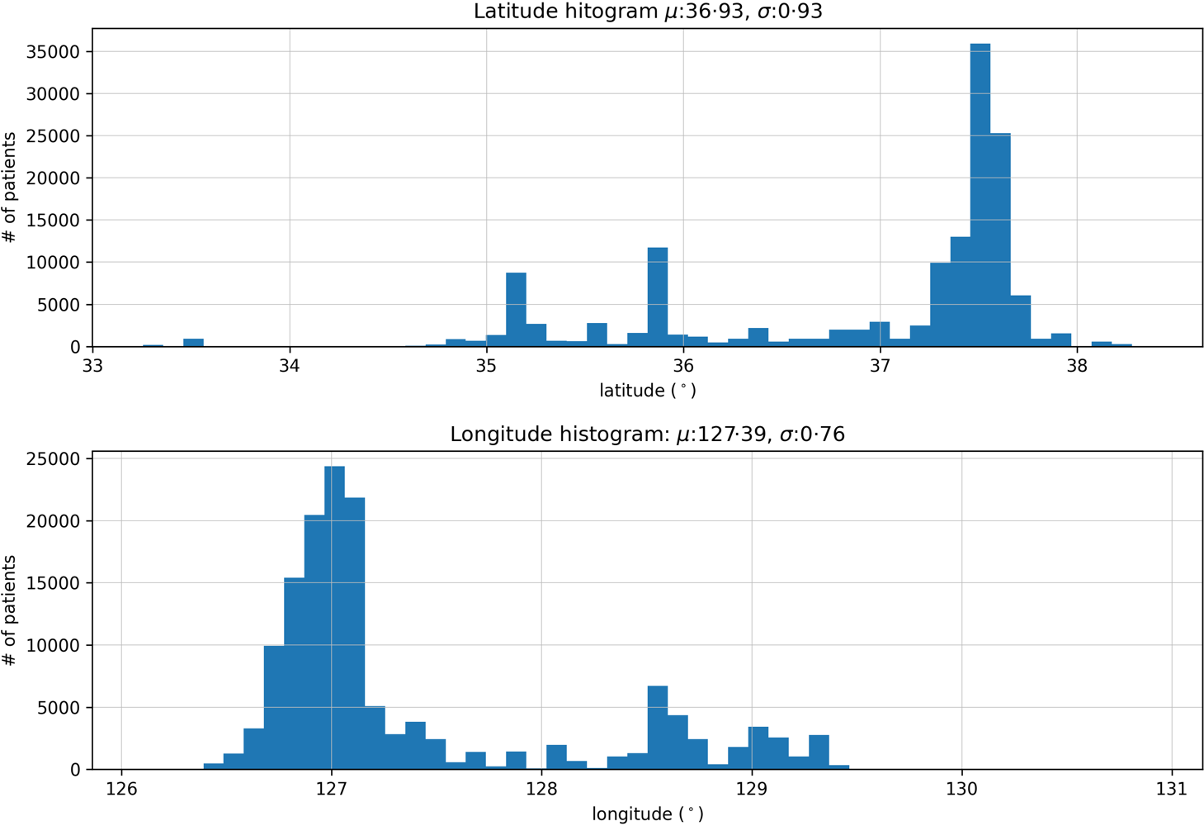}
    \caption{Histogram of patients' distribution by latitude (top) and longitude (bottom).}
    \label{fig:fig3}
\end{figure}

The distribution of longitude and latitude of the study participants is presented in Figure \ref{fig:fig3}. The $x$-axis represents the latitude-longitude coordinate, while the $y$-axis shows its number of patients. The $\mu$ and $\sigma$ in the title denote the mean and the standard deviation, respectively. Even though discrepancies between the actual area of residence and lat-long pair exist, they were ignored because such cases were rare.

\begin{figure}[!htb]
    \centering
    \includegraphics[width=.8\textwidth]{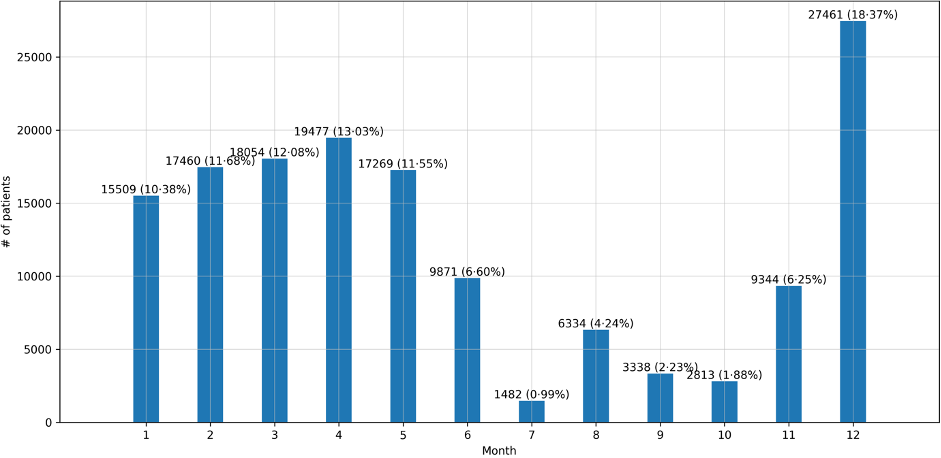}
    \caption{Cumulative number of confirmed cases per month.}
    \label{fig:fig4}
\end{figure}

The seasonality of the cumulative number of confirmed cases per month is presented in Figure \ref{fig:fig4}. The height of each bar represents the number of patients in that month. We marked the number of patients and their percentage (\%) at the top of the bar.

\begin{figure}[!htb]
    \centering
    \begin{subfigure}{.9\textwidth}
    \includegraphics[width=\linewidth]{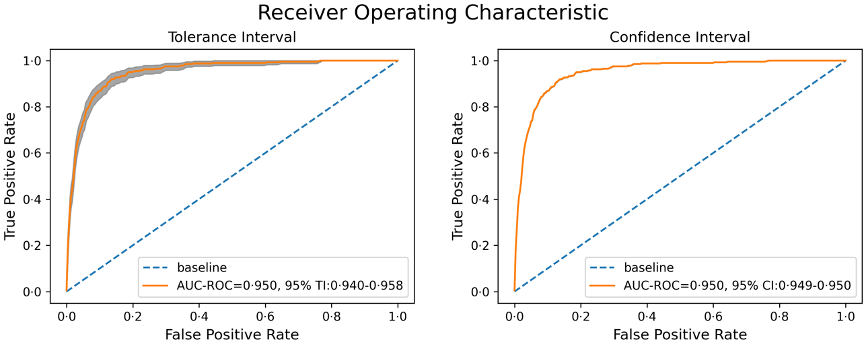}
    \caption{}
    \end{subfigure}
    \begin{subfigure}{.9\textwidth}
    \includegraphics[width=\linewidth]{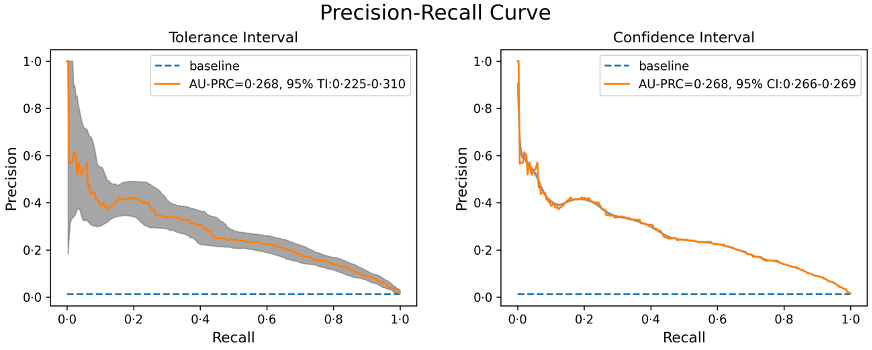}
    \caption{}
    \end{subfigure}
    \caption{(a) ROC Curve and (b) Precision-Recall Curve. The gray bands around the curves are pointwise 95\% TI and 95\% CI, which are derived by bootstrapping with 1,000 repetitions.}
    \label{fig:fig5}
\end{figure}

\noindent\textbf{Model performance} \smallskip\\
\indent The proposed model achieved an AUROC score of $0$·$950$ at a $95$\% Tolerance Interval (TI): $0$·$940$-$0$·$958$ and $95$\% Confidence Interval (CI): $0$·$949$-$0$·$950$, Youden’s index of $0$·$739$, F1-score of $0$·$861$, recall $0$·$807$, precision $0$·$923$, and specificity $0$·$933$. Since the size of the test set was 29,895, and there were 398 positives in the test set, the fraction of positives is $0$·$013$, which is the baseline for the AUPRC score. The model achieved an AUPRC score of $0$·$268$ (with $95$\% TI: $0$·$225$-$0$·$310$ and $95$\% CI: $0$·$266$-$0$·$269$), greatly outperforming the baseline score of $0$·$013$. The general ROC curve and PR curve are presented in Figure \ref{fig:fig5}.

We compared four different models with their performance presented in Table \ref{tab:tab4}. The XGB model achieved the highest scores with an AUROC of $0$·$950$ and AUPRC of $0$·$268$.

\begin{table}[!htb]\centering
\captionsetup{singlelinecheck= false, format=hang, justification=centering}
\caption{Performances of four different models.}
\begin{tabular}{|c|cccc|}
\hline
\multicolumn{1}{|l|}{} & XGBoost & LightGBM  & Random Forest & CatBoost \\ \hline
AUPRC                  & $0$·$268$   & $0$·$260$     & $0$·$240$         & $0$·$261$    \\
AUROC                  & $0$·$950$   & $0$·$943$     & $0$·$944$         & $0$·$947$    \\
Precision              & $0$·$923$   & $0$·$925$     & $0$·$978$         & $0$·$881$    \\
Recall                 & $0$·$807$   & $0$·$769$     & $0$·$025$         & $0$·$897$    \\
F1                     & $0$·$861$   & $0$·$840$     & $0$·$049$         & $0$·$889$    \\
Youden's Index         & $0$·$739$   & $0$·$707$     & $0$·$025$         & $0$·$776$    \\
Specificity            & $0$·$933$   & $0$·$938$     & $0$·$999$         & $0$·$879$    \\ \hline
\end{tabular}\label{tab:tab4}
\end{table}

\begin{figure}[!htb]
    \advance\leftskip0cm
    \includegraphics[width=.8\textwidth]{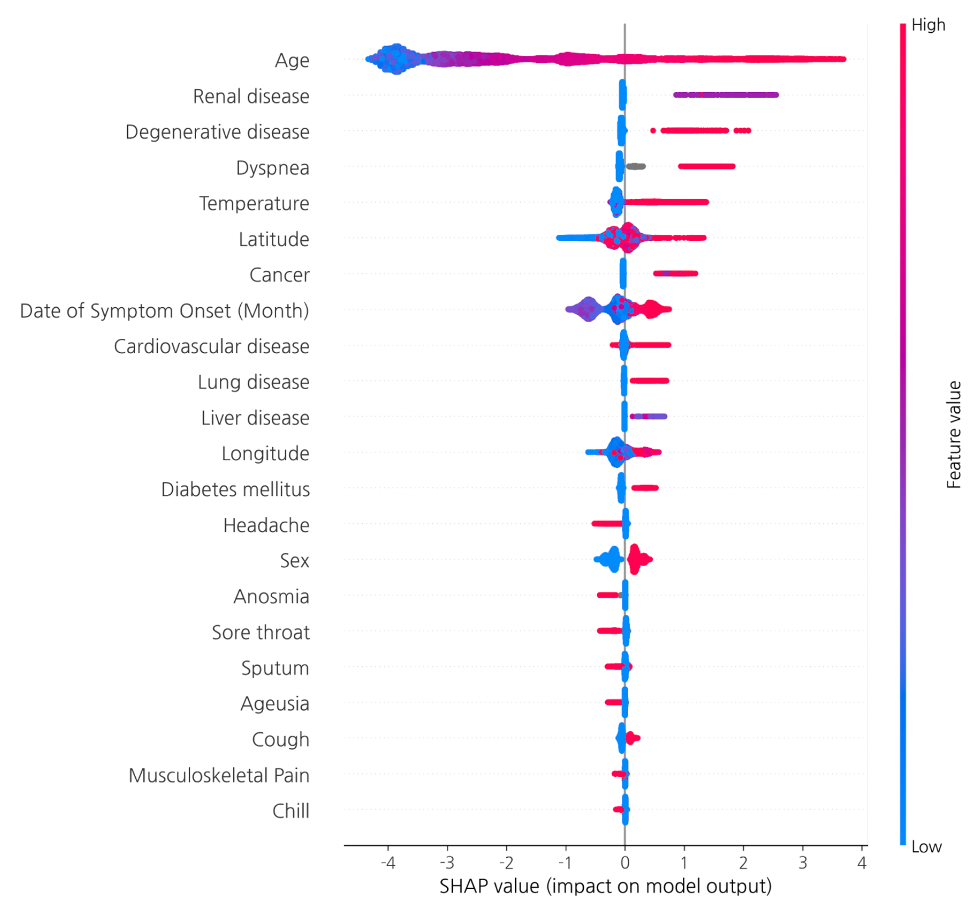}
    \caption{Feature importance plot.}
    \label{fig:fig6}
\end{figure}

\noindent\textbf{Explainability} \smallskip\\
\indent Feature importance was measured by SHAP, as presented in Figure \ref{fig:fig6}. Features in the plot are sorted in descending order by their maximum absolute values. A single dot on each row represents the explanation for each patient, and the original feature values are represented by their colors. The SHAP analysis proved age to be the most important relevant risk factor for mortality. Body temperature was also an important risk factor, as were previous diseases before COVID-19 infection, such as renal disease, degenerative disease, cancer, liver, cardiovascular, and lung disease. Among initial symptoms of patients, dyspnea was shown to be an important risk factor. Geographic information is also closely related to the mortality of COVID-19 patients. Higher longitude and latitude are related to high mortality. The Northeast region is covered with more mountains than the west or southern region in S. Korea while almost all large cities are located in the southern and western parts of the country. In terms of accessibility to acute care facilities, geographic location significantly affects the mortality of patients with acute respiratory diseases \cite{32parcha2021trends,33kishamawe2019trends}. Different weather according to location may also affect the severity of disease or mortality of the patients \cite{34sil2020does}.

\noindent\textbf{Cost-benefit analysis} \smallskip\\
\indent Decision curve analysis (DCA), as depicted in Supplementary Figure \ref{fig:fig7}, provides the range of threshold probabilities in which a prediction model shows the value and magnitude of benefit \cite{35vickers2006decision}. In the context of this research, the threshold can be used to decide whether a self-quarantined patient should be hospitalized or not. The threshold should be set depending on the medical and economic environment of the country in which the model is implemented. The DCA identified the optimal threshold range in which net benefit does not fall below zero. In our model, the optimal threshold for the DCA ranged from $0$ to $0$·$05$.

\begin{figure}[!htb]
    \centering
    \begin{subfigure}{.45\textwidth}
    \includegraphics[width=\linewidth, height=.2\textheight]{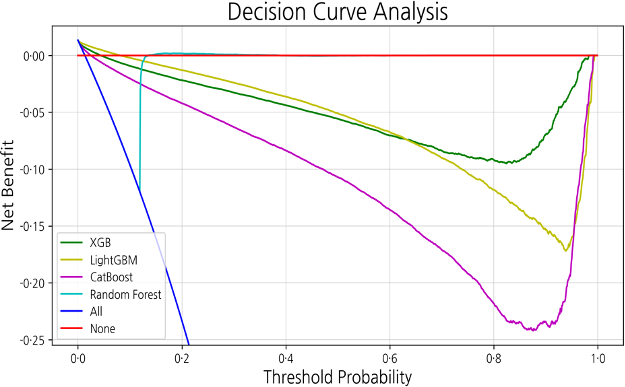}
    \end{subfigure}
    \begin{subfigure}{.45\textwidth}
    \includegraphics[width=\linewidth, height=.2\textheight]{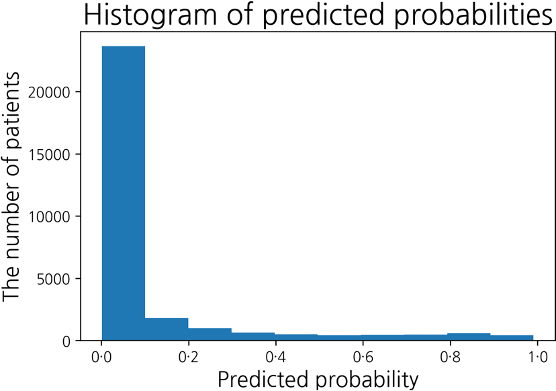}
    \end{subfigure}
    \caption{Decision curve analysis and the histogram of predicted probabilities of the XGB model.}
    \label{fig:fig7}
\end{figure}

\begin{figure}[!htb]
    \advance\leftskip2.5cm
    \includegraphics[width=.8\textwidth]{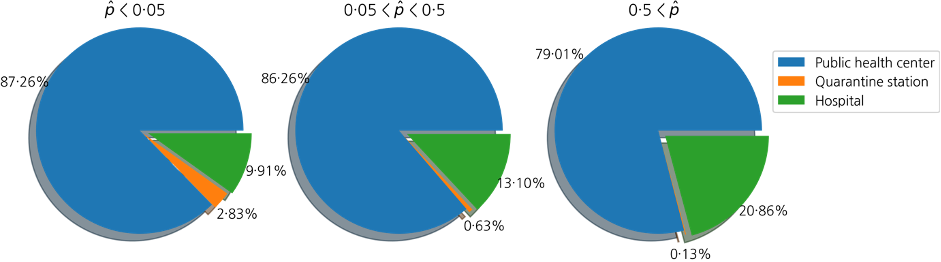}
    \caption{First-visit facility of COVID-19 patients according to the patients’ mortality probabilities.}
    \label{fig:fig8}
\end{figure}

We also investigated the types of medical institutions visited by patients according to their predicted mortality probabilities, as shown in Figure \ref{fig:fig8}. First, we divided the test set into three groups: patients with predicted mortality probabilities less than $0$·$05$, those between $0$·$05$ and $0$·$5$, and those greater than $0$·$5$. Then, we analyzed the types of medical institutions that the patients visited first for each group. Since public health centers are the first places where patients receive the PCR test in general, the proportion of public health centers among the medical institutions where patients get treated is great. However, the proportion of hospitals in the pie chart increases if the mortality rate of patients increases, which means more severely infected patients visited hospitals at first than those with less severe cases.

\section{Conclusion / Discussion}

In this research, we propose a machine learning model that predicts the prognosis of SARS-CoV-2 infected patients by obtaining 20 basic PGHD. The model was developed using the data of 149,471 patients from 1,382 designated COVID-19 screening centers. Thus, our model can be utilized globally for triaging confirmed patients of SARS-CoV-2 at the initial stage and monitoring hospitalized or quarantined patients daily.

The characteristics of SARS-CoV-2 and the related spectrum of signs and symptoms are the subjects of much ongoing research. Initial triage of the patients is crucial to prevent the shutdown of the entire medical system of a country. Thus, there have been many studies on developing patient triage algorithms using easily obtainable signs and symptoms. The model in this study provides a novel method integrating easily obtainable signs and symptoms, along with geographic and seasonal data that reflect characteristics of respiratory viruses, all from the nationwide multicenter database, including hospitalization and mortality data.

Accurate patient triage may lower the burden currently faced by health systems through facilitating optimized management of healthcare resources during future waves of the SARS-CoV-2 pandemic \cite{36hamid2020current}. This is especially important in developing countries with limited resources to maintain essential health services \cite{37world2020maintaining}.

While reviewing the existing research, we found that most of the previous studies utilized limited data. Furthermore, almost all of them utilized various input features that are not easy to measure. Compared to the previous studies, we adopted two types of demographic information, one geographic location, one sign, nine symptoms, and seven underlying diseases, which are easy-to-measure. Only body temperature and the nine symptoms are changeable during quarantine and hospitalization. Thus, patients can check the severity of the disease every day with the variable input features. The data for the research was collected from 1,382 designated COVID-19 screening centers in S. Korea, which means the developed model covered patients with variant clinical characteristics from all over the country. In addition, we adopted longitude and latitude in our model to reflect clinical characteristics of the acute respiratory virus by weather and accessibility of acute care facilities of each region.

Through the result of DCA, users can set a threshold for intervention such as transfer to a higher level of care or medical facility or a thorough examination by doctors. For example, if they are allowed to have a higher false-positive rate and want to screen necessary patients for intervention as much as possible, they can set the threshold near $0$. If they have to save hospital beds for severe patients when medical resources are depleting, they can set the value closer to $0$·$05$.

The SHAP analysis found patients with previous renal, degenerative, or cardiovascular diseases or cancer should be monitored thoroughly. In addition, body temperature and dyspnea should be considered the most important factors to assess aggravation of their health daily.

One of the main limitations of the study is that our model has not yet been extensively applied to the field. Therefore, we could not quantify how efficiently our model could lower the burden on the healthcare system. However, since our model has high performance and easily accessible, we expect to have positive results and leave this analysis as a future work.

In conclusion, we developed a model for predicting COVID-19 diagnosis by obtaining 20 basic PGHD based on nationwide multicenter data reported by KDCA. With the help of COVID-19 vaccination and medicine to be released soon, it will be more important to manage patients under quarantine at home or facility. Our framework can be implemented and utilized conveniently to triage patients with positive RT-PCR test results as well as enabling them to monitor themselves at home or a quarantine facility.


\section{Acknowledgment}

We thank the participants of the study.

\noindent\textbf{Authorship} \smallskip\\
\indent All named authors meet the International Committee of Medical Journal Editors (ICMJE) criteria for authorship for this article, take responsibility for the integrity of the work as a whole, and have given their approval for this version to be published.

\noindent\textbf{Author Contributions} \smallskip\\
\indent M.S.P. and H.J. analyzed data and created the models and drafted the entire manuscript as the first authors. H.L. contributed to the discussion of the results. S.Y.J. and H.J.H. revised the manuscript and supervised the entire process as the corresponding authors.

\noindent\textbf{Compliance with Ethics Guidelines} \smallskip\\
\indent This research was approved by the Institutional Review Board of Seoul National University Bundang Hospital (X-2110-717-902). An Informed consent form was unobtained due to the nature of retrospective studies. The study was performed in accordance with the Helsinki Declaration of 1964 and its later amendments.

\noindent\textbf{Funding} \smallskip\\
\indent This research was supported by the Seoul National University Bundang Hospital Research Fund (Grant \#14-2021-0041), the National Research Foundation of Korea Grant funded by the Korean government (NRF-2017R1E1A1A03070105 and NRF-2019R1A5A1028324), the ITRC (Information Technology Research Center) support program (IITP-2018-0-01441), the Institute for Information \& Communications Technology Promotion grant funded by the Korean government (Artificial Intelligence Graduate School Program [POSTECH]; \#2019-0-01906), and KAIST Stochastic Analysis and Application Research Center.

\noindent\textbf{Disclosures} \smallskip\\
\indent Min Sue Park, Hyeontae Jo, Haeun Lee, Se Young Jung and Hyung Ju Hwang have nothing to disclose.

\noindent\textbf{Data Availability} \smallskip\\
\indent Data is not available to the public due to the regulation of KDCA.

\bibliographystyle{unsrtnat}
\bibliography{references}  

\end{document}